\documentclass[10pt,twocolumn,letterpaper]{article}

\usepackage{iccv}              

\usepackage[accsupp]{axessibility}  

\usepackage{amssymb}
\usepackage{pifont}
\newcommand{\cmark}{\ding{51}}%
\newcommand{\xmark}{\ding{55}}%
\usepackage{multirow}
\usepackage{comment}
\usepackage{wrapfig,lipsum}
\usepackage{listings}
\usepackage{algorithm}
\usepackage{algpseudocode}
\usepackage{amsmath}
\usepackage{soul}
\usepackage{pifont}

\makeatletter
\def\@fnsymbol#1{%
   \ifcase#1\or
   \TextOrMath \textdagger \dagger\or
   \TextOrMath \textdaggerdbl \ddagger \or
   \TextOrMath \textsection  \mathsection\or
   \TextOrMath \textparagraph \mathparagraph\or
   \TextOrMath \textbardbl \|\or
   \TextOrMath {\textdagger\textdagger}{\dagger\dagger}\or
   \TextOrMath {\textdaggerdbl\textdaggerdbl}{\ddagger\ddagger}\else
   \@ctrerr \fi
}
\makeatother

\definecolor{iccvblue}{rgb}{0.21,0.49,0.74}
\usepackage[pagebackref,breaklinks,colorlinks,allcolors=iccvblue]{hyperref}

\def\paperID{13941} 
\def\confName{ICCV}
\def\confYear{2025}

\title{Adapting In-Domain Few-Shot Segmentation to New Domains \\ without Source Domain Retraining}

\author{
Qi Fan\textsuperscript{1}, 
Kaiqi Liu\textsuperscript{1}, 
Nian Liu\textsuperscript{2}\thanks{Corresponding author.}, 
Hisham Cholakkal\textsuperscript{2}, 
Rao Muhammad Anwer\textsuperscript{2}, 
Wenbin Li\textsuperscript{1}, 
Yang Gao\textsuperscript{1}
\\
\textsuperscript{1}Nanjing University\quad 
\textsuperscript{2}Mohamed bin Zayed University of Artificial Intelligence \\
}

\begin{document}
\maketitle
\begin{abstract}
Cross-domain few-shot segmentation (CD-FSS) aims to segment objects of novel classes in new domains, which is often challenging due to the diverse characteristics of target domains and the limited availability of support data.
Most CD-FSS methods redesign and retrain in-domain FSS models using abundant base data from the source domain, which are effective but costly to train.
To address these issues, we propose adapting informative model structures of the well-trained FSS model for target domains by learning domain characteristics from few-shot labeled support samples during inference, thereby eliminating the need for source domain retraining.
Specifically, we first adaptively identify domain-specific model structures by measuring parameter importance using a novel structure Fisher score in a data-dependent manner.
Then, we progressively train the selected informative model structures with hierarchically constructed training samples, progressing from fewer to more support shots.  
The resulting Informative Structure Adaptation (ISA) method effectively addresses domain shifts and equips existing well-trained in-domain FSS models with flexible adaptation capabilities for new domains, eliminating the need to redesign or retrain CD-FSS models on base data.
Extensive experiments validate the effectiveness of our method, demonstrating superior performance across multiple CD-FSS benchmarks.
Codes are at \url{https://github.com/fanq15/ISA}.

\end{abstract}

\section{Introduction}
\label{sec:intro}




Few-shot semantic segmentation (FSS) aims to segment novel classes using a limited number of support samples.
It typically trains a conventional support-query matching network to transfer class-agnostic patterns from extensive base data to novel classes.
Existing FSS methods~\citep{fan2022self,nguyen2019feature,lu2021simpler,zhang2019canet} have made significant progress on in-domain class generalization due to various well-designed model architectures, as well as matching and training techniques.



Despite their success, existing few-shot segmentation methods often struggle with domain shifts, particularly when test and training data come from different domain distributions.
This challenge underscores the significance of cross-domain few-shot segmentation (CD-FSS), which aims to generalize to new classes and unseen domains using minimal annotated data from the target domain.

\begin{figure}[!t]
  \centering
  \includegraphics[width=1.0\linewidth]{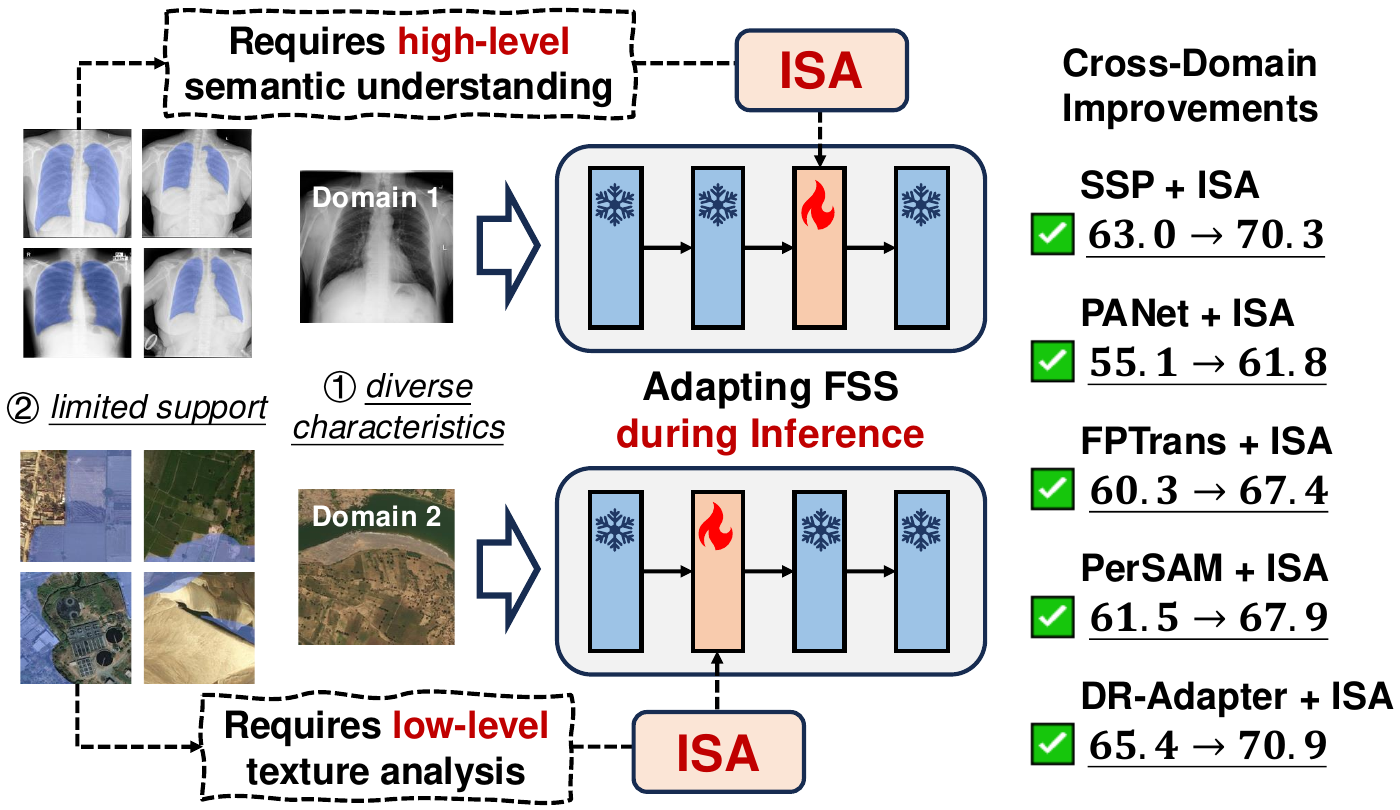}
  \caption{Cross-domain few-shot segmentation is usually challenging due to \ding{192} \ul{\it diverse characteristics of target domains} (\eg, requiring high- or low-level parsing) and the \ding{193} \ul{\it limited support data} (\eg, 1 to 5 shots). Our Informative Structure Adaptation (ISA) can adaptively identify and efficiently train domain-specific informative structures for the target domains {\bf during inference}. This process is applied to few-shot annotated support data during inference and can be directly integrated with various well-trained FSS methods~\cite{fan2022self,wang2019panet,zhang2022feature,zhang2023personalize,su2024domain} {\bf without source domain retraining}, leading to substantial improvements in CD-FSS.}
  \label{fig:teaser}
  \vspace{-0.15in}
\end{figure}

Existing CD-FSS methods~\citep{lei2022cross,su2024domain,wang2022remember,huang2023restnet} typically train models using abundant base data from the source domain with various domain-generalization techniques to address the domain shifts problem.
However, retraining on source domain is costly and fails to leverage powerful in-domain FSS models.
Besides, the issue arises because CD-FSS models are trained on limited domain data with frozen parameters, while the potential target domains can be diverse and arbitrary.
Therefore, it is necessary to adapt the well-trained in-domain FSS models to target domains during inference. 


Test-time training (TTT)~\citep{wang2020tent,sun2020test} effectively adapts models to target domains by learning from test data.
Although promising, most existing TTT methods adjust the same manually-selected trainable structures across different domains, disregarding the distinct domain gaps and characteristics of the target domains.
However, different domains, or even individual test images, exhibit distinct properties related to various model structures, such as specific model layers.
For example, the satellite images in DeepGlobe~\citep{demir2018deepglobe} dataset primarily rely on low-level texture analysis for parsing structured and detailed remote sensing areas.
In contrast, Chest X-ray images~\citep{candemir2013lung} typically require middle- and high-level semantic understanding to distinguish pneumonia-affected areas from normal lung regions in medical image analysis.
\ul{\it Thus, dynamically selecting trainable model structures is crucial for adapting to various target domains with distinct characteristics.}

To address these challenges, we propose a novel method, {\it Informative Structure Adaptation (ISA)}, specifically designed for cross-domain few-shot segmentation.
We explore methods for {\it identifying} and {\it adapting} domain-specific informative structures during inference by learning domain characteristics from few-shot labeled support samples.
Given the varying reliance on model structures across different domains~\citep{yosinski2014transferable,liang2019distant}, the {\it Informative Structure Identification (ISI)} module identifies domain-sensitive model structures by measuring parameter importance in a data-dependent manner.
First, we compute empirical Fisher information to reduce computational overhead, inspired by the parameter importance metric used in continual learning~\citep{kirkpatrick2017overcoming}.
Building on this, we propose a novel structure Fisher score to guide the identification of informative model structures.
This strategy optimizes model adaptation for varying domain characteristics and mitigates the risk of overfitting in few-shot scenarios.

\ul{\it Once informative model structures are obtained, optimizing trainable parameters becomes essential, particularly in few-shot scenarios.}
Conventional test-time training methods typically optimize the model using a single test image that shares the same class space as the training data.
In contrast, CD-FSS requires simultaneous generalization to new classes and unseen target domains, utilizing multiple labeled support images.
Therefore, it is essential to fully utilize few-shot support data to tackle the challenges of class and domain generalization.
We propose a novel {\it Progressive Structure Adaptation (PSA)} module that trains the model with hierarchically constructed training samples, progressing from fewer to more support shots.
Specifically, we initially create support-query training pairs by cyclically designating each support image as the pseudo query image.
Subsequently, we extend the training pairs by progressively increasing the number of support shots.
This strategy enables the model to gradually adapt to domain shifts and maximizes the use of limited support data during inference for the challenging CD-FSS.

Our approach, as shown in Figure~\ref{fig:teaser}, fundamentally differs from conventional few-shot segmentation and test-time training methods.
We leverage few-shot annotated support data to adaptively identify and progressively adapt informative model structures during test-time training for CD-FSS.
In contrast, most prevailing test-time training methods~\citep{wang2020tent,su2022revisiting,wang2022continual,su2024towards} manually define trainable model structures and directly train models on accessible test data, which are proven suboptimal by our empirical analysis.
Although PATNet~\citep{lei2022cross} introduces a test-time training method for CD-FSS, it is specifically designed to train fixed anchor layers and requires pre-training on the source domain.
Therefore, applying PATNet to other few-shot semantic segmentation methods is non-trivial.
In contrast, our method is model-agnostic, requires no additional learnable parameters, and can easily equip existing few-shot segmentation models with flexible adaptation capabilities for new domains, eliminating the need to redesign or retrain CD-FSS models on base data.
In summary, our key contributions include:
\begin{itemize}
    \item We propose, for the first time, a novel framework Informative Structure Adaptation (ISA)  for CD-FSS that generalizes well-trained in-domain few-shot segmentation models to new domains without source domain retraining.
    \item The Informative Structure Identification (ISI) module dynamically identifies domain-sensitive model structures in a data-dependent manner, while the Progressive Structure Adaptation (PSA) module progressively addresses domain shifts by adapting the model with an increasing number of support shots.
    \item Our ISA generalizes effectively across multiple unseen target domains and is remarkably simple. Extensive experiments and analysis validate the effectiveness of our proposed ISA method.
\end{itemize}

\section{Related Works}

\noindent{\bf Few-Shot Semantic Segmentation.}
Few-shot semantic segmentation, pioneered by Shaban et al.~\citep{shaban2017one}, aims to predict dense masks for objects of novel classes using only a limited number of labeled support images. 
The mainstream prototype-based methods \citep{dong2018few,li2021adaptive,wang2019panet} perform segmentation by measuring the similarity between the query features and representative support prototypes incorporating various improvements~\citep{siam2020weakly,liu2020dynamic,zhang2021self,zhuge2021deep}.
The affinity-based methods \citep{lu2021simpler,zhang2021few,peng2023hierarchical,min2021hypercorrelation,tian2020prior} establish detailed dense correspondence between query and support features through feature concatenation and leverage a learnable CNN or transformer module for segmentation prediction.
Recently, foundation models like SAM~\citep{kirillov2023segment} present a novel opportunity for few-shot segmentation~\citep{liu2023matcher,zhang2023personalize}, due to their remarkable transfer capability on tasks and data distributions beyond the training scope.
However, these methods do not consider the domain shifts problem, leading to poor generalization performance when encountering new domains during testing.

\noindent{\bf Cross-Domain Few-Shot Semantic Segmentation.}
Cross-domain few-shot semantic segmentation has recently received increasing attention. PATNet~\citep{lei2022cross} introduces a feature transformation layer that seamlessly maps query and support features across diverse domains into a unified feature space, effectively tackling the intra-domain knowledge preservation issue in CD-FSS. RD~\citep{wang2022remember} utilizes a memory bank to reinstill meta-knowledge from the source domain, thereby improving generalization performance in the target domain. Subsequently, DARNet~\citep{fan2023darnet} and RestNet~\citep{huang2023restnet} approach the problem from distinct perspectives, focusing on bridging domain gaps through dynamic adaptation refinement and knowledge transfer, respectively. Inspired by these pioneering efforts, PMNET~\citep{chen2024pixel} presents a comprehensive solution capable of addressing both in-domain and cross-domain FSS tasks concurrently by capturing pixel relations within each support-query pair.
Unlike previous CD-FSS approaches, our method effectively addresses domain shifts in CD-FSS during inference by adaptively identifying and gradually adapting informative model structures on few-shot annotated support samples, and is capable of seamlessly adapting current few-shot segmentation methods to address domain shifts problem.

\noindent{\bf Test-Time Training.}
Normally, once a well-trained model is deployed, it remains static without further alterations. 
In contrast, test-time training (TTT)~\citep{sun2020ttt} adapts models to the deployment scenario by leveraging unlabeled data available at test time.
The mainstream self-supervised learning-based methods~\citep{liu2021ttt++,wang2021tent,liang2020do,goyal2022test,gandelsman2022test} leverage the available unlabeled test data to facilitate model adaptation to the target domain using self-supervised learning techniques.
The feature alignment-based methods~\citep{su2022revisiting,jung2023cafa,wang2023feature} attempt to rectify the feature representations for the target domain.
Some works attempt to apply TTT to address the semantic segmentation problem.
For instance, MM-TTA~\citep{shin2022mmtta} utilizes multiple modalities to provide reciprocal TTT self-supervision for 3D semantic segmentation. Similarly, CD-TTA~\citep{song2022cdtta} explores domain-specific TTT for urban scene segmentation using an online clustering algorithm. OCL~\citep{zhang2024test} proposes an output contrastive loss to stabilize the TTT adaptation process for extreme class imbalance and complex decision spaces in semantic segmentation.
These methods typically adjust fixed hand-selected trainable structures on one single test image for different domains. In contrast, our method dynamically adapt domain-specific informative structures by learning domain characteristics from few-shot labeled support samples.


\section{Method}



Cross-Domain Few-Shot Segmentation (CD-FSS) aims to transfer knowledge learned from the source domain to new categories in unseen target domains using minimal annotated supports.
The model is typically trained on the source domain and then evaluated on target domains, ensuring no label space overlap between the source and target domains.	

\subsection{Baseline Methods}

\noindent{\bf Few-Shot Segmentation Model.} 
Mainstream few-shot segmentation model can be formulated as follows: The input support and query images $\{I_s, I_q\}$ are processed by a weight-shared backbone to extract features $\{\mathcal{F}_s, \mathcal{F}_q\}$:
\begin{equation}\label{feature extractor}
    \mathcal{F}_s = f(I_s; \theta), \mathcal{F}_q = f(I_q; \theta),
\end{equation}
where $f$ denotes the image encoder with parameters $\theta$.
Then, the support features $\mathcal{F}_s$ and groundtruth masks $\mathcal{M}_s$ are fed into the masked average pooling layer (MAP) to generate support prototypes $\mathcal{P}_s$. The prediction is made by measuring the cosine similarity between $\mathcal{P}_s$ and $\mathcal{F}_q$. 

\begin{algorithm*}[!t]
\caption{ Informative Structure Adaptation (ISA)}
\label{algorithm}
\begin{algorithmic}[1]
\State \textbf{Require:} $K$-shot support samples, and well-trained FSS model $\mathit{f}$.
\State Select trainable layer parameters $\theta_{\texttt{tr}}$ using {\bf ISI} module \Comment{See Section~\ref{section:3.2}}
\For {$n$ from $1$ to $K-1$} \Comment{\textbf{PSA with increasing support shots.} See Eqn.~\ref{parameter update progressively}}
    \State Extract features $\mathcal{F}$ for all samples using the updated model $\mathit{f}$ \Comment{See Eqn.~\ref{feature extractor}}
    \For {\textit{i} from $1$ to $K$} \Comment{\textbf{HTC with cyclic pseudo query}}
        \State Compute loss $\mathcal{L}_{\texttt{T}}^i$ on $\mathcal{F}$ for the $i$-th query with $n$ support samples \Comment{See Eqn.~\ref{TTTloss}}
        \State (Omit the computation on combinations of $n$ support samples for clarity)
    \EndFor
    \State Compute average loss $\mathcal{L}_{\texttt{PSA},n}$ for  support shots $n$
    \Comment{See Eqn.~\ref{HTCloss}}
    \State Back-propagation for model $\mathit{f}$ with $\mathcal{L}_{\texttt{PSA},n}$ 
    \State Update model
    $\mathit{f}$: $\theta_{\texttt{tr},n-1}^*$ $\rightarrow$ $\theta_{\texttt{tr},n}^{*}$
    \Comment{See Eqn.~\ref{HTCoptimize}}
\EndFor

\end{algorithmic}
\end{algorithm*}

\noindent{\bf Model Structure Adaptation Baseline.} 
Our model structure adaptation baseline for CD-FSS is derived from the Test-Time Training (TTT) method, which typically adapts the pre-trained source domain model during evaluation using the available test data.
In CD-FSS, we segment the unlabeled query image using the few-shot support set $S=\left\{\left(I_{s}^{i}, \mathcal{M}_{s}^{i}\right)\right\}_{i=1}^{K}$ containing $K$ support images with groundtruth masks.
We leverage the labeled support data to train the few-shot matching model for adapting model structures by constructing support-query pairs with mask labels.
Specifically, we randomly select one support data $S^i_{q}=(I_{s}^{i}, \mathcal{M}_{s}^{i})$ as the pseudo query data, and treat the remaining support samples as a new support set $S\setminus S^i_{q}$, creating a support-query training pair $(S \setminus S^i_{q}, S^i_{q})$.
Then, we extract support prototypes $\mathcal{P}_s^i$ and query features $\mathcal{F}_{q}^i$ for test-time training:
\begin{equation}\label{TTTloss}
    \mathcal{L}_{\texttt T}^i=BCE\left(\operatorname{cosine}\left(\mathcal{P}_{s}^i, \mathcal{F}_{q}^i\right), \mathcal{M}_{q}^i\right),
\end{equation}
where BCE is the binary cross entropy loss and $\mathcal{M}_{q}$ is the groundtruth mask of the pseudo query image.
Eventually, we train the model by optimizing the loss:
$
    \theta^{*}=\underset{\theta}{\arg \min } \mathcal{L}^{i}_{\texttt{T}} (\mathcal{P}_{s}^i,\mathcal{F}_{q}^i,\mathcal{M}_{q}^i;\theta),
$
where $\theta$ denotes the trainable parameters of the model and $\theta^{*}$ denotes the updated model parameters after training.
To prevent overfitting, we follow the common practice~\citep{boudiaf2021few,he2020momentum} to train only the final convolutional layer of the model during test-time training.




\subsection{Informative Structure Identification}
\label{section:3.2}

To adapt the model to varying domain characteristics, we first investigate {\it how to identify} domain-specific informative structures from few-shot labeled support data during inference, rather than manually defining trainable model layers.

\noindent{\bf Fisher Information Matrix} (FIM) can evaluate the significance of parameters concerning a specific task and data distribution.
Given a model with parameters $\theta$, input $x_i$, output $y_i$ and output probability $p_{\theta}(y_i | x_i)$, the FIM can be computed as
$F_\theta = \mathbb{E}_{x \sim p(x)} \left[ \mathbb{E}_{y \sim p_\theta(y|x)} \left( \frac{\partial \log p_\theta(y|x)}{\partial \theta} \right) \left( \frac{\partial \log p_\theta(y|x)}{\partial \theta} \right)^{\top} \right]$. The matrix \( F_\theta \in \mathbb{R}^{|\theta| \times |\theta|} \) can alternatively be understood as representing the covariance of the gradients of the log likelihood with respect to the parameters \( \theta \).


\noindent{\bf Empirical Fisher Information.} However, directly computing the Fisher Information Matrix for the pre-trained CD-FSS model involves significant computational overhead due to the \( |\theta| \times |\theta| \) scale.
Thus, we simplify FIM for CD-FSS, inspired by the parameter importance metric used in continual learning~\citep{kirkpatrick2017overcoming}. Specifically, we concentrate on the support samples \( K \) in the target domain and utilize the diagonal elements of the ``Empirical Fisher'' to evaluate the importance of the pre-trained model parameters for cross-domain tasks. Specifically, for the $l$-th convolutional layer, we derive the empirical Fisher information of its $u$-th trainable parameters as $F_{\theta_{l,u}} = \frac{1}{|K|} \sum_{j=1}^{|K|} \left( \frac{\partial \log p_{\theta}(y_j | x_j)}{\partial \theta_{l,u}} \right)^2$.
Correspondingly, a relatively large value of \( F_{\theta_{l,u}} \) indicates that the parameter \( \theta_{l,u} \) is crucial for the cross-domain task.

\noindent{\bf Structure Fisher Score.} Now, we can compute the empirical Fisher information for all parameters of the model based on labeled support samples.
We observe that the empirical Fisher information is typically distributed sparsely throughout the model, with many low-value entries in each convolutional layer.
Directly fine-tuning the most sensitive unstructured parameters may lack the representational capacity to handle severe domain shifts.
Therefore, we propose identifying informative model structures, \ie, convolutional layers, for subsequent model adaptation.
Specifically, we compute the maximum absolute value of empirical Fisher information across all $U$ trainable parameters within the $l$-th layer as its structure fisher score $F^{*}_{\theta_l}$:
\begin{equation}\label{gradient_representation}
    F^{*}_{\theta_l} = \max \left(|F_{\theta_{l,1}}|,|F_{\theta_{l,2}}|,\dots,|F_{\theta_{l,u}}|,\dots,|F_{\theta_{l,U}}|\right).
\end{equation}


Model layers with higher structure Fisher scores are typically more important for model training~\citep{liu2021sparse} because of their greater contribution to the optimization process.
Updating only the informative model structures preserves the model's ability to fit few-shot data and regularizes training to mitigate the risk of overfitting.
Therefore, we select the model layer with the highest structure Fisher score and update its parameters $\theta_{\texttt{tr}}$ for model structure adaptation during inference, while freezing all other parameters to minimize the risk of overfitting in few-shot scenarios:
\begin{equation}\label{trainable layer chosen}
    \theta_{\texttt{tr}}=\theta_{l^{*}}, {\mathrm{where}} \ 
    l^{*}=\underset{l}{\arg \max }\{F^{*}_{\theta_l}\}.
\end{equation}
Further discussions on the usage of structure Fisher Score for CD-FSS is included in the supplementary material.

\subsection{Progressive Structure Adaptation} 

After identifying informative model structures, it is essential to optimize trainable parameters during inference by fully utilizing few-shot support data to tackle the challenges of CD-FSS.
Therefore, we propose a Progressive Structure Adaptation (PSA) module to assist the model gradually address domain shifts by adapting informative model structures on hierarchically constructed training samples, progressing from fewer to more support shots.



\begin{table*}[!t]
\centering
\caption{Quantitative comparison results on the CD-FSS benchmark. The models are trained on the Pascal VOC source domain dataset and evaluated on four datasets exhibiting distinct domain shifts. The best results are highlighted with {\bf bold}. The $\dagger$ means our reproduced results. The $\ddagger$ means using the ViT-base backbone. 
All the in-domain FSS methods are directly applied to CD-FSS without source domain retraining (\xmark), thus suffering from the domain shifts problem. All other CD-FSS methods requires source domain retraining (\cmark) their models on the source domain to enhance domain generalization. Our method can be directly integrated with the well-trained FSS method (\eg, SSP) without source domain retraining (\xmark).}
  \resizebox{0.9\linewidth}{!}{
  \renewcommand{\tabcolsep}{1.4mm}
  \vspace{-0.25in}

\begin{tabular}{c|c|c|cccccccc|cc}
\toprule
\multirow{2}{*}{Methods} & \multirow{2}{*}{Publication} & \multirow{2}{*}{Retraining} & \multicolumn{2}{c}{Deepglobe} & \multicolumn{2}{c}{ISIC} & \multicolumn{2}{c}{Chest X-ray} & \multicolumn{2}{c}{FSS-1000} & \multicolumn{2}{|c}{mIoU} \\
\cmidrule(lr){4-5} \cmidrule(lr){6-7} \cmidrule(lr){8-9} \cmidrule(lr){10-11} \cmidrule(lr){12-13}
& & & 1-shot & 5-shot & 1-shot & 5-shot & 1-shot & 5-shot & 1-shot & 5-shot & 1-shot & 5-shot \\
\midrule
PGNet~\citep{zhang2019pyramid} & ICCV 2019 & \xmark &  10.7 & 12.4 & 21.9 & 21.3 & 34.0 & 23.0 & 62.4 & 62.7 & 32.2 & 31.1 \\
PANet~\citep{wang2019panet} & ICCV 2019 & \xmark & 36.6 & 45.4 & 25.3 & 34.0 & 57.8 & 69.3 & 69.2 & 71.7 & 47.2 & 55.1 \\
CaNet~\citep{zhang2019canet} & CVPR 2019 & \xmark & 22.3 & 23.1 & 25.2 & 28.2 & 28.4 & 28.6 & 70.7 & 72.0 & 36.6 & 38.0 \\
RPMMs~\citep{yang2020prototype} & ECCV 2020 & \xmark & 13.0 & 13.5 & 18.0 & 20.0 & 30.1 & 30.8 & 65.1 & 67.1 & 31.6 & 32.9 \\
PFENet~\citep{tian2020prior} & TPAMI 2020 & \xmark & 16.9 & 18.0 & 23.5 & 23.8 & 27.2 & 27.6 & 70.9 & 70.5 & 34.6 & 35.0 \\
RePRI~\citep{boudiaf2021few} & CVPR 2021 & \xmark & 25.0 & 27.4 & 23.3 & 26.2 & 65.1 & 65.5 & 71.0 & 74.2 & 46.1 & 48.3 \\
HSNet~\citep{min2021hypercorrelation} & ICCV 2021 & \xmark & 29.7 & 35.1 & 31.2 & 35.1 & 51.9 & 54.4 & 77.5 & 81.0 & 47.6 & 51.4 \\
SSP$^\dagger$~\citep{fan2022self} & ECCV 2022 & \xmark & 42.3 & 50.4 & 33.0 & 47.0 & 74.9 & 75.5 & 77.1 & 79.1 & 56.8 & 63.0 \\
ABCDFSS~\citep{herzog2024adapt} & CVPR 2024 & \xmark & 42.6 & 49.0 & 45.7 & 53.3 & 79.8 & 81.4 & 74.6 & 76.2 & 60.7 & 65.0 \\
\midrule
PATNet~\citep{lei2022cross} & ECCV 2022 & \cmark & 37.9 & 43.0 & 41.2 & 53.6 & 66.6 & 70.2 & 78.6 & 81.2 & 56.1 & 62.0 \\
APSeg$^\ddagger$~\citep{he2024apseg} & CVPR 2024 & \cmark & 35.9 & 40.0 & 45.4 & 54.0 & {\bf 84.1} & 84.5  & 79.7 & 81.9 &  {\bf 61.3} & 65.1 \\
DR-Adapter~\citep{su2024domain} & CVPR 2024 & \cmark & 41.3 & 50.1 & 40.8 & 48.9 & 82.4 & 82.3 & 79.1 & 80.4 & 60.9 & 65.4 \\
DMTNet~\cite{chen2024cross} & IJCAI 2024 & \cmark & 40.1 & 51.2 & 43.6 & 52.3 & 73.7 & 77.3 & 81.5 & 83.3 & 59.7 & 66.0 \\
\midrule
Ours & - & \xmark & {\bf 44.3} & {\bf 52.7} & 37.2 & {\bf 56.1} & 83.4 & {\bf 86.3} & 78.8 & 86.0 & 60.9 & {\bf 70.3} \\
\bottomrule
\end{tabular}}
\label{table:first}
\vspace{-0.15in}
\end{table*}

\noindent{\bf Hierarchical Training Sample Construction.} We begin by enhancing the utilization of few-shot support data, cyclically designating each support image as a pseudo query image to generate multiple support-query training pairs $\{(S \setminus S^i_{q}, S^i_{q})\}_{i=1}^{K}$.
To conserve computational resources, we first extract features from all support images, then compute the losses for each constructed training pair based on these features, and finally perform back-propagation to update model parameters using the averaged loss $\frac{1}{K} \sum_{i=1}^K \mathcal{L}_{\texttt{T}}^i$ across all training pairs.
Next, we vary the number of support shots from $1$ to $K-1$ to construct hierarchical training pairs.
Specifically, for each support shot number $n$ ($n \leq K-1$), we construct training pairs containing $n$ support samples.
Thus, we progressively increase the number of support samples to construct training pairs, optimizing the utilization of few-shot support data and enabling the model to gradually handle domain shifts during inference.

\noindent{\bf Progressive Structure Adaptation.} 
To better address domain shifts in cross-domain tasks during test-time training, we propose a Progressive Structure Adaptation (PSA) module that trains the model using HTC-constructed hierarchical training pairs.
The PSA method progressively trains the model with an increasing number of support shots from $1$ to $K-1$, gradually reducing the domain gap.
Specifically, for the support shot number $n$, the training loss is:
\begin{equation}\label{HTCloss}
    \mathcal{L}_{\texttt{PSA},n}=\frac{1}{K \cdot S_n}\sum_{i=1}^{K}\sum_{s_n} B C E\left(\operatorname{cosine}\left(\mathcal{P}_{s_n}^i, \mathcal{F}_{q}^i\right), \mathcal{M}_{q}^i\right),
\end{equation}

where $s_n$ denotes the $s_n$-th combination of $n$ support samples, enumerated from the $S\setminus S^i_{q}$ support set, with a total of $S_n$ combinations.
The model parameters are updated by optimizing the loss:
\begin{equation}\label{HTCoptimize}
    \theta_{n}^{*}=\underset{\theta_{n-1}^{*}}{\arg \min }\mathcal{L}_{\texttt{PSA},n} (\mathcal{P}_{s_n}^i,\mathcal{F}_{q}^i,\mathcal{M}_{q}^i;\theta_{n-1}^{*}),
\end{equation}
where $\theta_{n-1}^{*}$ and $\theta_{n}^{*}$ denote the updated model parameters trained with HTC-constructed training samples with $n-1$ and $n$ support shots, respectively. Specifically, when $n=1$, the $\theta_{n-1}^{*}$ represents the original parameters $\theta$ of the FSS model.

Consequently, we progressively update the model parameters by gradually increasing the number of support shots from $1$ to $K-1$, as follows:
$
\theta_1^{*}\rightarrow\theta_2^{*}\rightarrow\dots\rightarrow\theta_n^{*}\rightarrow\dots \rightarrow\theta_{K-1}^{*}.
$
This approach optimizes the use of limited support data to progressively mitigate domain shifts during inference on the support set. Hence, we term it as Progressive Structure Adaptation (PSA) module.


\subsection{Informative Structure Adaptation}


We incorporate the proposed ISI and PSA modules into the Informative Structure Adaptation (ISA) method, as shown in Algorithm~\ref{algorithm}.
Specifically, we first use ISI to select the trainable parameters $\theta_{\texttt{tr}}$ for test-time training.
Then, we use PSA with hierarchically constructed training pairs to train the selected model parameters by gradually increasing the number of support shots $n$ from $1$ to $K-1$:
\begin{equation}\label{parameter update progressively}
\theta_{{\texttt{tr}},1}^{*}\rightarrow\theta_{{\texttt{tr}},2}^{*}\rightarrow\dots\rightarrow\theta_{{\texttt{tr}},n}^{*}\rightarrow\dots \theta_{{\texttt{tr}},K-1}^{*}.
\end{equation}

Notably, unlike conventional online TTT settings, our method isolates model training among testing episodes, thereby safeguarding against data leakage and ensuring fidelity to the few-shot setting.

Our ISA method can also be applied to transformer-based methods~\cite{zhang2023personalize,zhang2022feature} by adaptively selecting and tuning the fully-connected layers of the backbone during inference.

\section{Experiments}

\begin{figure*}[!t]
  \centering
  \includegraphics[width=0.8\linewidth]{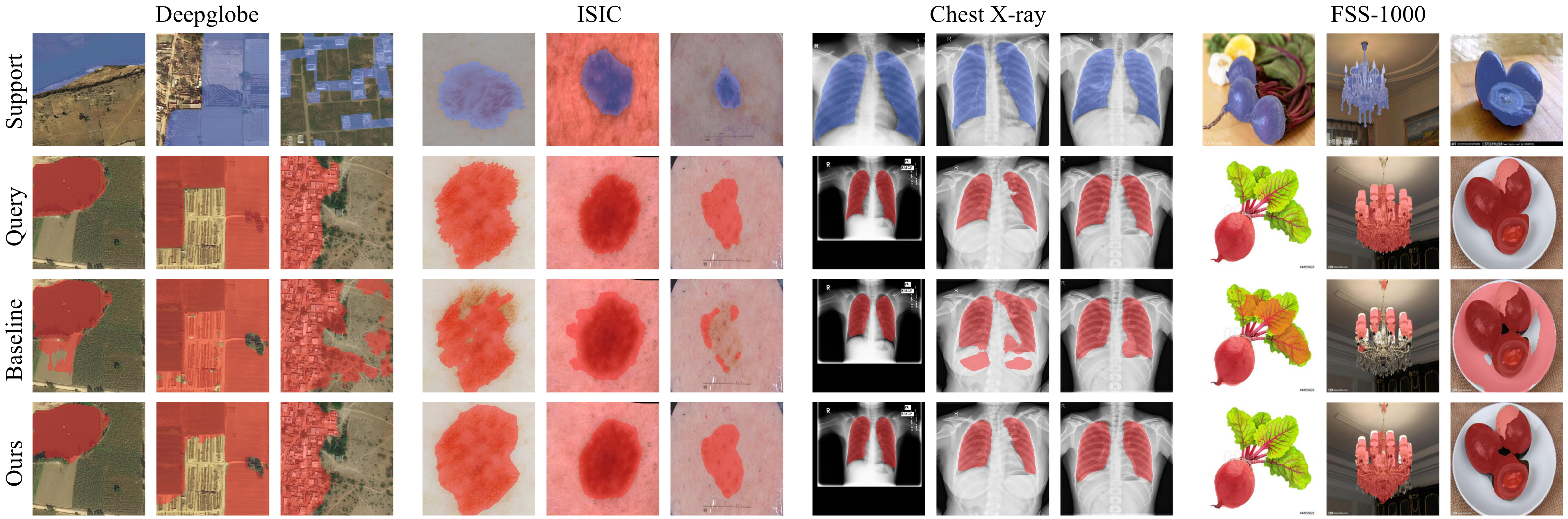}
  \vspace{-0.1in}
  \caption{Qualitative comparisons between our method and the baseline model in the 1-way 5-shot setting across four target domain datasets. We show only one support image for clarity.}
  \label{fig:vis}
  \vspace{-0.15in}
\end{figure*}

We adopt the popular few-shot semantic segmentation model SSP~\citep{fan2022self} as our baseline, trained on Pascal VOC~\citep{everingham2010pascal} source domain dataset. 
We directly apply our method to the public released, well-trained SSP model with a ResNet-50~\citep{he2016deep} backbone,  without any re-training on the source domain dataset.
For test-time training, we use the SGD optimizer with a learning rate of 1e-3 and one training iteration to update the trainable model parameters.
Following previous works~\citep{lei2022cross,su2024domain}, we evaluate all methods on four datasets with distinct domain shifts: Deepglobe~\citep{demir2018deepglobe} for satellite images with seven categories, ISIC2018~\citep{codella2019skin,tschandl2018ham10000} for medical images with three types of skin lesions, Chest X-Ray~\citep{candemir2013lung,jaeger2013automatic} for medical screening images, and FSS-1000~\citep{li2020fss} for 1000-class daily objects.
The input images are resized to $400 \times 400$ pixels.
In the 1-shot setting, we apply data augmentation to support images to generate two additional support images for test-time training. 
We use the mean Intersection-over-Union (mIoU) for evaluation.
All experiments are conducted on a Tesla V100 GPU.



\begin{table}[!t]
  \caption{Quantitative comparison results on SUIM dataset, where models are trained on Pascal VOC.}
  \label{table:suim}
  \small
  \centering
  \vspace{-0.1in}
  \renewcommand{\tabcolsep}{1.0mm}
  \resizebox{\columnwidth}{!}{ 
    \begin{tabular}{c|cccccccc}
      \toprule
      & ASGNet & HSNet & SCL & RD & DAM & MMT & DR-Adapter & Ours \\
      \midrule
      mIoU & 31.9 & 28.8 & 31.8 & 34.7 & 34.8 & 35.9 & 40.3 & {\bf 44.1} \\
      \bottomrule
    \end{tabular}
  }
  \vspace{-0.05in}
\end{table}

\begin{table}[!t]
  \caption{Results of applying our method to FSS/CD-FSS models with ResNet and Transformer backbones. ``In Domain?'' indicates whether the model is designed for in-domain FSS. ``w/ Ours'' denotes that the model is equipped with our ISA method. ``$\Delta$'' represents the performance improvement. All models are trained on Pascal VOC (except for PerSAM), and evaluated on four CD-FSS datasets. The average 5-shot mIoU performance is reported.}
  \label{table:generalize}
  \vspace{-0.1in}

  \centering
  \renewcommand{\tabcolsep}{1.0mm}
  \resizebox{\columnwidth}{!}{ 
    \begin{tabular}{c|cc|cc|c}
      \toprule
      FSS Model & Backbone & In Domain? & Original & w/ Ours & {\bf $\Delta$} \\
      \midrule
      SSP~\cite{fan2022self}  & ResNet-50 & \cmark & 63.0 & 70.3 & {\bf +7.3} \\
      PANet~\cite{wang2019panet} & ResNet-50 & \cmark & 55.1 & 61.8 & {\bf +6.7} \\
      FPTrans~\cite{zhang2022feature} & Transformer & \cmark & 60.3 & 67.4 & {\bf +7.1} \\
      PerSAM~\cite{zhang2023personalize} & Transformer & \cmark & 61.5 & 67.9 & {\bf +6.4} \\
      DR-Adapter~\cite{su2024domain} & ResNet-50 & \xmark & 65.4 & 70.9 
 & {\bf +5.5} \\
      \bottomrule
    \end{tabular}
  }
  \vspace{-0.15in}
\end{table}

\begin{figure*}[!t]
  \centering
  \includegraphics[width=0.8\linewidth]{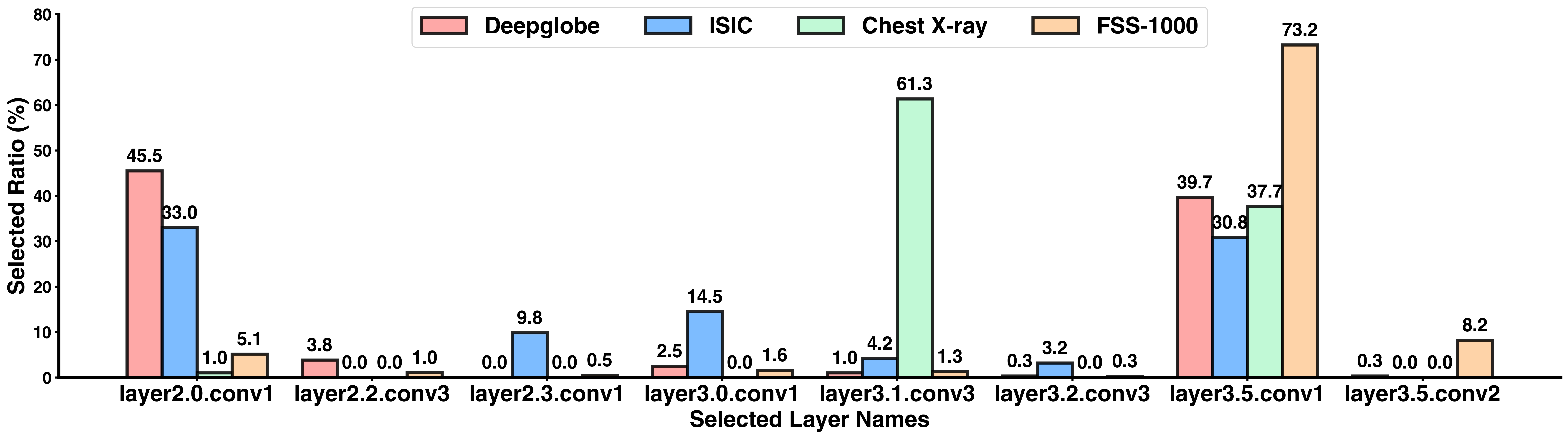}
    \vspace{-0.1in}

  \caption{Selected trainable layer distribution of the informative structure identification (ISI) module. The “selected ratio” denotes the frequency with which each layer is selected across the entire dataset.}
  \label{fig:selected}
  \vspace{-0.15in}
\end{figure*}

\subsection{Comparison with State-of-the-Arts}

In Table~\ref{table:first}, we compare our method with existing cross-domain few-shot semantic segmentation methods.
Our method substantially outperforms the baseline method SSP~\citep{fan2022self}, with a 4.1/7.3 mIoU average improvement in the 1-shot/5-shot settings across all datasets.
Additionally, our method surpasses previous CD-FSS SOTA methods, ABCDFSS~\citep{herzog2024adapt}, APSeg~\citep{he2024apseg} 
DR-Adapter~\citep{su2024domain} and DMTNet~\cite{chen2024cross}, by a large margin in the 5-shot setting. APSeg achieves slightly better performance than our method (61.3 $v.s.$ 60.9), primarily because their ViT backbone is more powerful than our ResNet-50 backbone.
Note that most other CD-FSS methods (except for ABCDFSS) {\bf require extensive source domain retraining, typically dozens of hours}, on the source domain dataset to learn transferable, domain-agnostic features for domain generalization.
In contrast, our method can effectively and efficiently adapt existing well-trained FSS models for segmenting objects of novel classes under domain shifts {\bf without any source domain retraining}.

To further validate the effectiveness of our method, we follow RD~\citep{wang2022remember} to evaluate our method on the SUIM dataset. 
All models are trained on Pascal VOC dataset and evaluated on SUIM~\citep{islam2020semantic} dataset. Table~\ref{table:suim} shows that our method improves the SOTA performance from 40.3 to 44.1 mIoU, beating other popular methods, including ASGNet~\citep{li2021adaptive}, HSNet~\citep{min2021hypercorrelation}, SCL~\citep{zhang2021self}, RD~\citep{wang2022remember}, DAM~\citep{chen2024dense}, MMT~\citep{wang2023mmt} and DR-Adapter~\citep{su2024domain}.

Recently, IFA~\citep{nie2024cross} sets a new SOTA on CD-FSS benchmarks, but they adopt a distinct evaluation protocol. For a fair comparison, we adopt their code and evaluation protocol to implement and evaluate our method.
When combined with their baseline model, our method achieves 72.8 mIoU, surpassing their reported 71.4 mIoU.

\noindent{\bf Qualitative Comparisons.} Figure~\ref{fig:vis} shows qualitative result comparisons between our method and the strong baseline SSP models. 
By applying our ISA method, we can substantially improve the segmentation quality of the FSS method for novel objects in distinct domains.

\noindent{\bf Generalized to Other Methods.} 
Our ISA method is general and can be applied to other FSS methods to address cross-domain few-shot semantic segmentation.
As shown in Table~\ref{table:generalize}, when equipped with our ISA method, two {\it ResNet-based} in-domain FSS methods, SSP~\cite{fan2022self} and PANet~\citep{wang2019panet} achieve substantially better performance on CD-FSS.
Our method can improve {\it transformer-based} in-domain FSS method FPTrans~\citep{zhang2022feature}, boosting the performance to 67.4 mIoU for CD-FSS.
Our method can also adapt the powerful {\it SAM-based} method PerSAM~\citep{zhang2023personalize} for CD-FSS, achieving remarkable 67.9 mIoU.
Our method can further improve existing {\it CD-FSS methods}, as evidenced by the 5.5 mIoU improvement on DR-Adapter~\citep{su2024domain} when combined with our ISA method.

\begin{table}[!t]
  \caption{Comparisons with more related methods.}
  \label{table:more}
  \centering
  \vspace{-0.1in}
    \small
  \renewcommand{\tabcolsep}{0.6mm}
  \resizebox{\columnwidth}{!}{ 
    \begin{tabular}{c|ccccc}
      \toprule
       & \multicolumn{3}{c|}{DG-based} & \multicolumn{2}{c}{TTT-based}  \\
      \cmidrule(lr){2-4} \cmidrule(lr){5-6} 
       & Mixstyle~\cite{zhou2021domain} & DSU~\cite{li2022uncertainty} & NP~\cite{fan2022towards} & TTT~\cite{sun2020test} & Tent~\cite{wang2020tent}  \\
      mIoU & 63.8 & 64.2 & 64.5 & 63.1 & 63.3 \\
      \midrule
        \midrule
       & \multicolumn{2}{c|}{PEFT-based} & \multicolumn{2}{c|}{SAM-based} & -  \\
      \cmidrule(lr){2-3} \cmidrule(lr){4-5} 
       & Adapter~\cite{chen2022adaptformer} & BitFit~\cite{zaken2022bitfit} & PerSAM~\cite{zhang2023personalize} & Matcher~\cite{liu2023matcher} & Ours  \\
      mIoU & 66.4 & 65.9 & 61.5 & 62.1 & 70.3  \\
      \bottomrule
    \end{tabular}
  }
  \vspace{-0.2in}
\end{table}

\noindent{\bf Comparison with More Related Methods.}
Table~\ref{table:more} compares our method with domain generalization (Mixstyle~\citep{zhou2021domain}, DSU~\citep{li2022uncertainty}, and NP~\citep{fan2022towards}), test-time training (TTT~\citep{sun2020test} and Tent~\citep{wang2020tent}), PEFT-based methods (Adapter~\cite{chen2022adaptformer} and BitFit~\cite{zaken2022bitfit}, PEFT for Parameter-Efficient Fine-Tuning), and SAM-based methods (PerSAM~\citep{zhang2023personalize} and Matcher~\citep{liu2023matcher}) to further demonstrate the superiority of our approach.
Our method significantly outperforms other related methods.

\subsection{Informative Structure Adaptation Analysis}
\label{section:4.3}

We conduct extensive experiments to understand our informative structure adaptation method. 
All experiments are performed in the 5-shot setting, focusing solely on the target module of the full ISA.

\noindent{\bf Module Ablation.}
In Table~\ref{table:ablation}, the simple model structure adaptation (MSA) module improves the performance by 1.6 mIoU, thanks to the model adaptation for diverse target domains. 
The ISI module improves segmentation performance on all target domains, due to the identified informative structures for model adaptation to varying domain characteristics.
The PSA module boosts the performance to 67.6 mIoU, attributing to its progressive training strategy to gradually solve domain shifts and maximal exploitation of the few-shot data.
Integrating all modules, our ISA method significantly improves the performance from 63.0 to 70.3 mIoU on the strong baseline model.

\begin{table}[!t]
  \caption{Results of ablation studies for our method. ``MSA'' denotes the model structure adaptation baseline, ``ISI'' denotes the informative structure identification module, and ``PSA'' denotes the progressive structure adaptation module.}
  \label{table:ablation}
  \centering
  \vspace{-0.1in}
  \renewcommand{\tabcolsep}{1.0mm}

  \resizebox{\columnwidth}{!}{ 
    \begin{tabular}{ccc|cccc|c}
      \toprule
      MSA & ISI & PSA & Deepglobe & ISIC & Chest X-ray & FSS-1000 & mIoU  \\
      \midrule
      & & & 50.4 & 47.0 & 75.5 & 79.1 & 63.0  \\
      \cmark & & & 50.9 & 48.4 & 80.8 & 78.2 & 64.6  \\
      \cmark & \cmark & & 51.5 & 50.6 & 81.7 & 82.1 & 66.5 \\
      \cmark & & \cmark & {\bf 53.2} & 50.8 & 84.2 & 82.0 & 67.6\\
      \cmark & \cmark & \cmark & 52.7 & {\bf 56.1} & {\bf 86.3} & {\bf 86.0} & {\bf 70.3}\\
      \bottomrule
    \end{tabular}
  }
  \vspace{-0.2in}
\end{table}

\begin{figure*}[!t]
  \centering
  \includegraphics[width=0.8\linewidth]{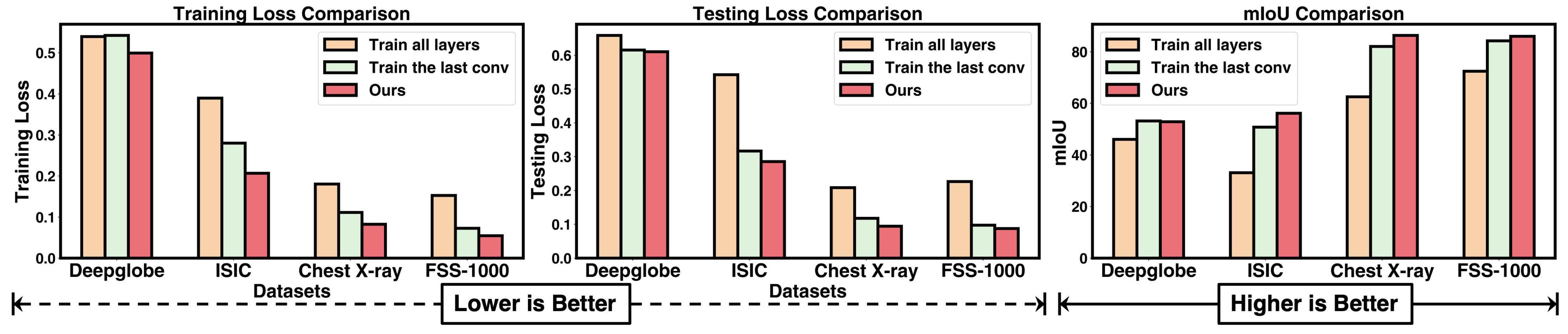}
    \vspace{-0.1in}

  \caption{Comparisons on training loss, testing loss and mIoU for various model training strategies.}
  \label{fig:loss}
  \vspace{-0.15in}
\end{figure*}

\noindent{\bf Informative Structure Identification Mechanism.}
Figure~\ref{fig:selected} summarizes the trainable layer distribution selected by ISI for various datasets. 
The ISI-selected trainable layers vary significantly depending on the properties of each dataset.
For example, the DeepGlobe and ISIC datasets both require reliable low-level texture analysis for accurate segmentation, thus guiding the model to select more low-level trainable layers, such as ``\texttt{layer2.0.conv1}''.
The FSS-1000 dataset requires high-level semantic understanding for segmenting various common objects in context, guiding the model to primarily train the high-level layers, such as ``\texttt{layer3.5.conv1}''.
Figure~\ref{fig:loss} compares the training loss, testing loss and mIoU for various model training strategies.
When training all model layers, both the training loss and testing loss are significantly high across all datasets, indicating inferior generalization ability.
Training only the last convolutional layer mitigates the overfitting problem.
Our ISI strategy further addresses the overfitting problem, evidenced by the lowest training and testing losses, achieving the best generalization performance.
This analysis validates the effectiveness and working mechanism of our selective self-guiding mechanism in addressing overfitting for test-time training in CD-FSS.

\noindent{\bf Informative Structure Identification Hyperparameters.}
Table~\ref{table:ssg} summarizes the model performance under different ISI settings. 
Compared to training only the last convolutional layer, our ISI substantially improves performance from 67.6 to 70.3 mIoU.
By increasing the number of trainable layers, performance can be further boosted to 71.0 mIoU when ISI selects three trainable layers.
We find that structure Fisher score are distributed sparsely, with many low-value scores in the convolutional parameters.
Thus, the average structure Fisher score of each convolutional layer cannot represent their importance, resulting in inferior segmentation performance.
In contrast, we compute the largest structure Fisher scores of each convolutional layer for trainable layer selection. 
We also experiment with computing the top-k largest structure Fisher scores for selecting trainable layers and achieve consistently good performance.

\begin{table}[!t]
  \caption{Results of various hyperparameteres for informative structure identification (ISI) module.}
  \label{table:ssg}
  \centering
    \vspace{-0.1in}

  \renewcommand{\tabcolsep}{1.3mm}
  \resizebox{\linewidth}{!}{
  \begin{tabular}{c|cccccc}
    \toprule
    & Manual & \multicolumn{5}{c}{ISI with different \# of trainable layers} \\
    \cmidrule(lr){2-2} \cmidrule(lr){3-7}
    & last conv & 1 layer & 2 layers & 3 layers & 4 layers & 5 layers \\
    mIoU & 67.6 & 70.3 & 70.8 & 71.0 & 70.7 & 70.6 \\
    \midrule
    \midrule
    & \multicolumn{6}{c}{ISI with different structure Fisher scores} \\
    \cmidrule(lr){2-7}
    & top1 & top3 & top5 & top10 & top20 & mean  \\
    mIoU & 70.3 & 70.2 & 70.3 & 70.3 & 70.2 & 67.8 \\
    \bottomrule
  \end{tabular}}
  \vspace{-0.1in}
\end{table}

\begin{table}[!t]
  \caption{Results of using various training strategies in the progressive structure adaptation (PSA) module. The ``1/2/3/4'' denotes training models using HTC-constructed samples with 1/2/3/4 support shots. The $\rightarrow$ denotes the sequential training procedure.}
  \label{table:psg}
    \vspace{-0.1in}

  \centering
  \renewcommand{\tabcolsep}{2mm}
  \footnotesize
  \resizebox{\linewidth}{!}{
  \begin{tabular}{c|cccc|ccc}
    \toprule
    & 1 & 2 & 3 & 4 & 1$\rightarrow$4 & 2$\rightarrow$4 & 3$\rightarrow$4 \\
    mIoU & 67.0 & 67.3 & 67.5 & 67.8 & 69.7 & 69.8 & 69.6 \\
    \bottomrule
  \end{tabular}}
\renewcommand{\tabcolsep}{2.5mm}
  \resizebox{\linewidth}{!}{
  \begin{tabular}{c|ccc|c}
    \toprule
    & 1$\rightarrow$2$\rightarrow$4 & 1$\rightarrow$3$\rightarrow$4 & 2$\rightarrow$3$\rightarrow$4 & 1$\rightarrow$2$\rightarrow$3$\rightarrow$4 \\
  
    mIoU & 70.0 & 70.1 & 70.1 & 70.3 \\
    \bottomrule
  \end{tabular}}
  \vspace{-0.2in}

\end{table}

\noindent{\bf Progressive Structure Adaptation Mechanism.} 
Our PSA module gradually addresses domain shifts using HTC-constructed training samples, progressing from fewer to more support images.
As shown in Table~\ref{table:psg}, when directly trained with the 4-shot training pairs, the model performs worse than our PSA-trained model, with a 2.5 mIoU performance drop.
By adding one intermediate training step ($1 \rightarrow 4$, $2 \rightarrow 4$, or $3 \rightarrow 4$), the performance drop is significantly reduced to 0.5-0.7 mIoU.
Adding more intermediate training steps further improves generalization performance.
These results validate the importance of progressive training in gradually addressing domain shifts.
Notably, our PSA does not require additional data and maximizes the use of limited support data to construct hierarchical training pairs for progressive self-guiding training.

\begin{table}[!t]
  \caption{Results of using various support shots for the baseline model and our method.}
  \label{table:shot}
  \vspace{-0.1in}
  \centering
  \renewcommand{\tabcolsep}{0.8mm}
  \resizebox{1.0\linewidth}{!}{
  \begin{tabular}{c|cccccccc}
    \toprule
    & 1 & 3-shot & 5-shot & 8-shot & 10-shot & 15-shot & 20-shot & 30-shot \\
    \midrule
    Baseline & 56.8 & 61.0 & 63.0 & 63.4 &  64.0 & 65.0 & 65.1 & 65.2  \\
    Ours & 60.9 & 67.1 & 70.3 & 71.0 & 71.9 & 73.0 & 74.0 & 74.6 \\
    \bottomrule
  \end{tabular}}
    \vspace{-0.2in}

\end{table}

\noindent{\bf Benefits from More Supports.}
Table~\ref{table:shot} shows that existing the TTT-free FSS method encounters performance saturation when the support data reaches 15 shots.
In contrast, our ISA method benefits from more support shots, reaching 74.6 mIoU with 30-shot supports.

\noindent{\bf Discussions on Foundation Model-based Methods.}
Foundation models, such as SAM~\citep{kirillov2023segment} and CLIP~\citep{radford2021learning}, are typically trained on large-scale web-collected data, resulting in excellent generalization on natural images.
However, due to data domain limitations, they often underperform in unseen domains like medical images, remote sensing images, or industrial images. Additionally, foundation models are typically built on large backbone models, leading to computation, deployment, and storage challenges.
In contrast, our method is specifically designed to address generalization to novel classes and unseen domains, featuring lightweight computation, a simple network architecture, few model parameters, and easy deployment. Our method is general and can flexibly equip foundation models to address domain shifts, evidenced in Table~\ref{table:generalize}.

\section{Conclusion}

We propose, for the first time, a novel framework Informative Structure Adaptation (ISA) that generalizes well-trained in-domain few-shot segmentation models to new domains without source domain retraining. Our Informative Structure Identification (ISI) adaptively identifies domain-specific model structures by measuring parameter importance with a novel structure Fisher score.
Furthermore, we propose the Progressive Structure Adaptation (PSA) to progressively adapt the selected informative model structures during inference, utilizing hierarchically constructed training samples with an increasing number of support shots.
The ISA method combines these strategies to effectively address domain shifts in CD-FSS, and equips existing few-shot segmentation models with flexible adaptation capabilities for new domains. 

\section*{Acknowledgements}

This work is supported in part by the National Natural Science Foundation of China (62192783, 62276128, 62406140), Young Elite Scientists Sponsorship Program by China Association for Science and Technology (2023QNRC001), the Key Research and Development Program of Jiangsu Province under Grant (BE2023019) and Jiangsu Natural Science Foundation under Grant (BK20221441, BK20241200)

{
    \small
    \bibliographystyle{ieeenat_fullname}
    \bibliography{main}
}

\section{Discussion on Fisher Information for ISA}

We provide a detailed explanation of the rationale for using Fisher information (FI) as a criterion for selecting and tuning layers and parameters in our ISA method.
We hope this discussion helps explain why Fisher information is particularly suitable for CD-FSS tasks.

\subsection{Theoretical Justification}

When considering the importance of parameters, we generally aim to understand how much changing a parameter will impact the model's output. Let us denote \( p_\theta(y | x) \) as the output distribution over \( y \) produced by a model with parameter vector \( \theta \in \mathbb{R}^{|\theta|} \), given input \( x \).

One way to measure the impact of a change in parameters on the model's prediction is by computing the Kullback-Leibler divergence (KL divergence) between the output distributions:
\begin{equation}
\text{D}_{\text{KL}}(p_\theta(y | x) || p_{\theta + \delta}(y | x)),
\end{equation}
where \( \delta \in \mathbb{R}^{|\theta|} \) represents a small perturbation. It can be shown that as \( \delta \to 0 \), we can perform a second-order Taylor expansion at \( \theta \), yielding:
\begin{equation}
\mathbb{E}_x \left[ \text{D}_{\text{KL}}(p_\theta(y | x) || p_{\theta + \delta}(y | x)) \right] = \delta^T F_\theta \delta + O(\delta^3),
\end{equation}
where \( F_\theta \in \mathbb{R}^{|\theta| \times |\theta|} \) is the Fisher information matrix, which is defined as:
\begin{equation}
F_\theta = \mathbb{E}_{x \sim p(x)} \mathbb{E}_{y \sim p_\theta(y | x)} \left[ \nabla_\theta \log p_\theta(y | x) \nabla_\theta \log p_\theta(y | x)^T \right].
\end{equation}

Due to the large size of \( F_\theta \), it is computationally intractable to compute it directly. As a result, we typically ignore the interactions between different parameters, and focus on the importance of each individual parameter in isolation. Therefore, following prior work, we approximate \( F_\theta \) by considering only its diagonal, which can be represented as a vector in \( \mathbb{R}^{|\theta|} \).

Thus, we obtain the following approximation for the Fisher information:
\begin{equation}
\hat{F}_\theta = \frac{1}{N} \sum_{i=1}^N \mathbb{E}_{y \sim p_\theta(y | x_i)} \left[ \left( \nabla_\theta \log p_\theta(y | x_i) \right)^2 \right],
\end{equation}
where \( \hat{F}_\theta \in \mathbb{R}^{|\theta|} \).

Although the Fisher information requires taking an expectation over the output distribution, in a supervised learning setting, we can leverage the availability of the ground-truth label \( y_i \) for each sample \( x_i \). This allows us to replace the expectation \( \mathbb{E}_{y \sim p_\theta(y | x_i)} \left[ \left( \nabla_\theta \log p_\theta(y | x_i) \right)^2 \right] \) in Eq. (1.4) with the squared gradient corresponding to the true label, \( \left( \nabla_\theta \log p_\theta(y_i | x_i) \right)^2 \).

This approach, known as "Empirical Fisher," provides a straightforward method for estimating the parameter importance heuristically. It can be computed as follows:
\begin{equation}
\hat{F}_\theta \approx \frac{1}{N} \sum_{i=1}^N \left( \nabla_\theta \log p_\theta(y_i | x_i) \right)^2,
\end{equation}
where \( p_\theta(y_i | x_i) \) denotes the output probability of \( y_i \) given input \( x_i \) and parameters \( \theta \), \( N \) is the number of samples, and \( \hat{F}_\theta \in \mathbb{R}^{|\theta|} \).

\subsection{Intuitive Interpretation}

Each element of the approximated Fisher Information vector \( \hat{F}_\theta \)  represents the expected squared gradient of the log-likelihood function with respect to a given parameter.
If a particular parameter strongly influences the model's output, its corresponding entry in \( \hat{F}_\theta \) will be large. Therefore, we can reasonably interpret \( \hat{F}_\theta \) as an approximation of the importance of each parameter in determining the model's output.

We hope this discussion provides a better understanding of the theoretical foundation and the intuitive interpretation of using Fisher information for selecting and tuning parameters.

\subsection{More Discussions}

The \textit{primary advantage of Fisher Information} over other methods~\cite{chen2018coupled,soen2021variance} lies in its principled, information-theoretic approach to quantifying the importance of model parameters. Unlike heuristic methods, which rely on rules of thumb or assumptions about parameter relevance, Fisher Information provides a mathematically rigorous measure of how sensitive the model’s output is to parameter changes. This makes FI particularly effective in settings with limited data, such as CD-FSS, where traditional methods like heuristic or gradient-based approaches may fail due to instability or high noise levels. For instance, \textit{heuristic methods} tend to be inconsistent in few-shot learning settings, and \textit{gradient-based methods}, while effective in large-data settings, can be noisy and less reliable in few-shot regimes due to data scarcity. On the other hand, \textit{Fisher Information} provides a more stable and consistent criterion for identifying parameter importance, which helps prioritize the parameters most critical to the model’s performance.

In our work, we focus on \textit{model structure adaptation}, where we leverage Fisher Information to identify and fine-tune only the most critical layers of the model, instead of fine-tuning all parameters. This approach is based on our observation that Fisher Information is sparsely distributed across the model, with many convolutional layers exhibiting low-value entries. To address this, we introduce \textit{structure Fisher scores}, which compute the maximum Fisher Information across all parameters within each layer. Layers with higher structure Fisher scores are typically more influential in the optimization process and, thus, more important for model adaptation. By focusing on the most informative layers, we ensure that fine-tuning efforts are more efficient, preserving the model’s ability to generalize while reducing the risk of overfitting in few-shot learning scenarios.

\textit{Why is Fisher Information particularly suitable for solving the problem of CD-FSS?}  
The key challenge in \textit{cross-domain few-shot segmentation} is the risk of overfitting, due to the limited number of training samples from the target domain. Fisher Information is particularly suited for this problem because it helps identify the model layers that are most crucial for generalization across domains. By targeting only the layers with the highest Fisher scores for adaptation, we can reduce the likelihood of overfitting, ensuring that the model focuses on the most important features for domain transfer. This targeted adaptation prevents unnecessary fine-tuning of less important parameters, which could introduce noise or instability, especially with limited data. Consequently, the model can better generalize to unseen domains without compromising performance.

In summary, \textit{Fisher Information} offers a more stable, principled, and efficient approach to parameter selection and model adaptation compared to traditional heuristic or gradient-based methods. It is particularly well-suited for \textit{cross-domain few-shot segmentation}, as it enables selective and targeted adaptation, helping to mitigate overfitting while enhancing generalization across domains.


\section{Limitations}
Although the proposed method demonstrates substantial performance improvements, it has inherent limitations that should be considered.
The additional modules introduced in the approach add complexity, potentially reducing processing speed due to extra model forwarding steps and increased computational overhead.
As a result, while the method excels in performance, it may not be suitable for applications that require high-speed processing, such as real-time systems or tasks with strict time constraints.

Despite these challenges, the proposed method is well-suited for performance-demanding applications where accuracy takes precedence over speed.
Future work could explore optimizations to balance the trade-off between performance and computational efficiency, possibly through more efficient model design or advanced hardware acceleration techniques.
These improvements could make the method more versatile for a broader range of practical applications.

We further propose a fast ISA method based on our experimental analysis.
Specifically, we replace ISI by directly selecting the ``\texttt{layer3.5.conv1}'' and the last convolutional layers as the trainable layers.
Additionally, we replace PSA with a ($2 \rightarrow 3 \rightarrow 4$)-based progressive training strategy, and randomly select only one training pair for each pseudo query data.
The proposed fast ISA method achieves 70.0 mIoU, with a considerable improvement on the running speed and a marginal performance drop compared with the original ISA method.

\end{document}